\pdfoutput=1

\documentclass[11pt]{article}

\usepackage{coling}

\usepackage{times}
\usepackage{latexsym}
\usepackage{booktabs}
\usepackage{xcolor}
\usepackage{graphicx}
\usepackage{float}
\usepackage{hyperref}
\usepackage{lipsum}

\usepackage[T1]{fontenc}

\usepackage[utf8]{inputenc}

\usepackage{microtype}

\usepackage{inconsolata}

\usepackage{graphicx}

\newcommand\blfootnote[1]{%
  \begingroup
  \renewcommand\thefootnote{}\footnote{#1}%
  \addtocounter{footnote}{-1}%
  \endgroup
}

\title{Measuring Contextual Informativeness in Child-Directed Text}

\author{Maria Valentini*$^\nabla$\quad~ Téa Wright*$^{\nabla\heartsuit}$\quad~ Ali Marashian$^\nabla$\quad~ Jennifer Weber$^\nabla$\\ \textbf{Eliana Colunga$^\nabla$\quad~ Katharina von der Wense$^{\nabla\diamondsuit}$} \\   
$^\nabla$University of Colorado Boulder \\   
$^\diamondsuit$Johannes Gutenberg University Mainz\\ 
$^\heartsuit$University of California Berkeley \\
{\tt \{first.last\}@colorado.edu}}

\begin{document}
\maketitle
\blfootnote{* denotes equal contribution.}

\begin{abstract}

To address an important gap in creating children's stories for vocabulary enrichment, we investigate the automatic evaluation of how well stories convey the semantics of target vocabulary words, a task with substantial implications for generating educational content. We motivate this task, which we call \textit{measuring contextual informativeness in children's stories}, and provide a formal task definition as well as a dataset for the task. We further propose a method for automating the task using a large language model (LLM). Our experiments show that our approach reaches a Spearman correlation of 0.4983 with human judgments of informativeness, while the strongest baseline only obtains a correlation of 0.3534. An additional analysis shows that the LLM-based approach is able to generalize to measuring contextual informativeness in adult-directed text, on which it also outperforms all baselines.  

\end{abstract}

\section{Introduction}
\label{sec:introduction}
Recent advances in natural language processing (NLP) have put the fully automated generation of children's stories within reach \cite{valentini}. Automatically-generated stories can be used for targeted vocabulary interventions for preschoolers when centered around desirable target words. As early vocabulary size is strongly correlated with reading ability in elementary school \cite{Walker_Greenwood_Hart_Carta_1994} and future academic success \cite{Brysbaert_Stevens_Mandera_Keuleers_2016}, such scalable interventions will contribute to leveling out existing inequalities.

\begin{figure}[th]
\setlength{\fboxsep}{0pt}%
\includegraphics[width=\columnwidth,keepaspectratio]{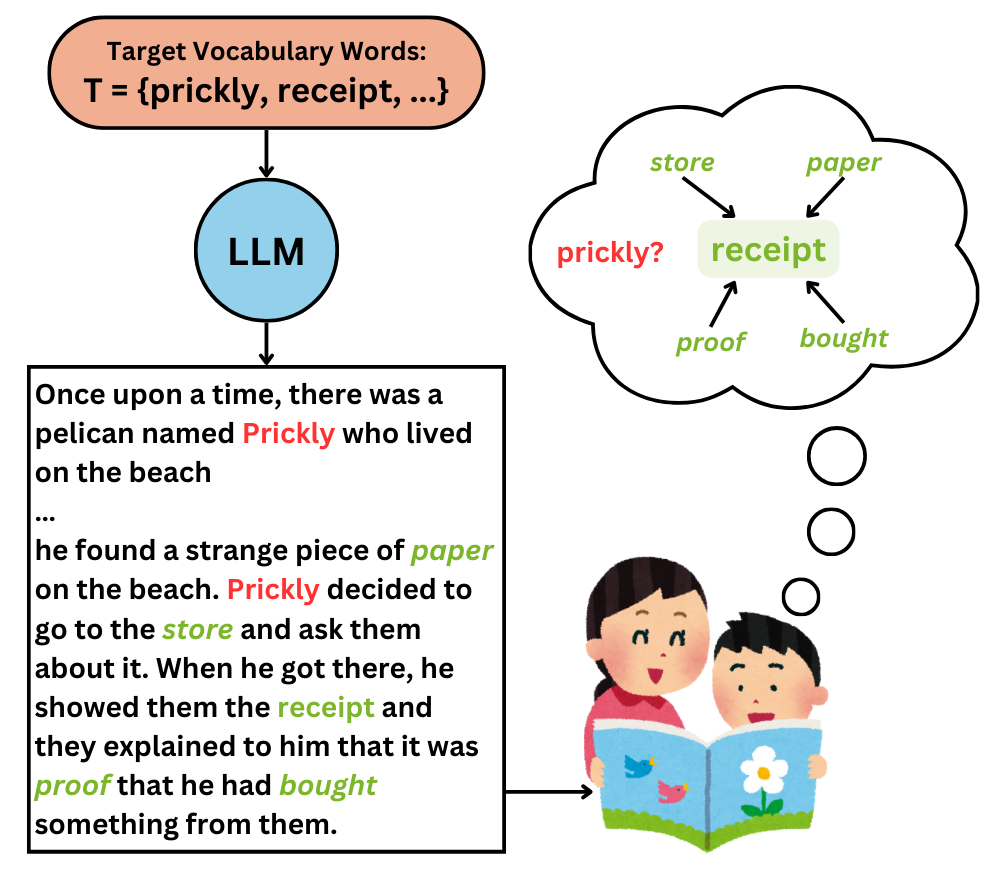}
    \caption{An example of an LLM-generated story providing an uninformative context for the word "prickly" and an informative context for the word "receipt." Italicized words represent helpful context terms in the passage corresponding to the target word of the same color.}
    \label{top_figure}
\end{figure}

Approximately 3,000 words are acquired each year in early childhood, primarily through incidental learning during reading \cite{Nagy_Anderson_1984}. However, just including target words in stories might not be enough for effective vocabulary enrichment: the amount of semantic information about a word in a story can vary widely. This issue is exasperated in stories generated by large language models (LLMs) when target words are often used in uninformative and misleading contexts; see Figure \ref{top_figure}.  

Automatically quantifying the amount of information about a word provided by a given story can help streamline the selection of effective stories for vocabulary learning and improve story generation models to support this purpose. 
With these benefits in mind, we introduce the task of \textit{measuring contextual informativeness in children's stories} and create a dataset for evaluation.

We further propose the use of Gemini \cite{Gemini} for this task with respect to a set of target words. We compare its performance with that of another proposed RoBERTa \cite{Liu2019RoBERTaAR}-based model, as well as multiple baselines. We find that, on the dataset we introduce, Gemini obtains a Spearman's $\rho$ value of 0.4983, while the RoBERTa-based model reaches 0.4601 and the strongest baseline only reaches 0.3534. We also show that our model generalizes to other domains, outperforming other approaches to measuring contextual informativeness in adult-directed text.

To summarize, we make the following contributions: (1) we propose the task of measuring contextual informativeness in children's stories; (2) we introduce a dataset for the task, which consists of automatically generated children's stories that have been annotated for the amount of contextual support they provide for target words; (3) we propose a method for the task and show that it outperforms multiple baselines; and (4) we demonstrate that our method generalizes for adult-directed text. Our dataset is available at \hyperlink{https://github.com/mariavale/contextual_inform}{https://github.com/mariavale/contextual$\_$inform}.

\section{Related Work}

\paragraph{In-context Vocabulary Learning}
As mentioned in Section \ref{sec:introduction}, research shows that the majority of new words are learned incidentally through reading in L1 learners \cite{Nagy_Anderson_1984}. As such, many modern vocabulary intervention approaches focus on in-context learning. 
Studies such as \citet{webb} provide evidence that 
more contextual clues about target words can lead to better learning outcomes. In addition, previous work stresses the importance of vocabulary intervention early in a child's life and the correlation of early vocabulary size with future academic success \cite{Walker_Greenwood_Hart_Carta_1994, Duff_Tomblin_Catts_2015, Brysbaert_Stevens_Mandera_Keuleers_2016}.

\paragraph{Cloze Task}
The cloze task \cite{taylor1953cloze} is designed to assess lexical and contextual understanding by removing words from a text, requiring participants to fill in the blanks with the missing words. Since its establishment, there has been disagreement about what the task truly measures. The evaluation of the task typically only allows one correct answer, raising concerns about how accurately it measures comprehension \cite{rapaport2005defense}. Despite this limitation, most experts agree it is indicative of understanding local vocabulary and semantic information \cite{Gellert_Elbro_2013, carlisle2004assessment}.
Many current language models pretrain on a masked language modeling objective, a form of the cloze task. 
Previous research has established that, for one such model, RoBERTa, prediction ability is correlated with human uncertainty \cite{jacobs-etal-2022-masked}. 

\paragraph{Learning Unknown Word Representations}
Previous work on learning representations of nonce and unknown words gives insight into how models may narrow down semantic space based on context. Nonce2Vec learns embeddings for unknown words from context and achieves high performance on a definitions dataset, but does not perform well with naturally occurring language. The authors hypothesize that adjusting risks taken during learning based on the informativeness of a context would improve results for naturally occurring language \cite{herbelot2017high}. \citet{schick2019learning} utilize two approaches --- (a) the surface-form representation (subword n-grams) and (b) learning an embedding from its context --- increasing performance compared to using either of the two approaches alone.

\paragraph{Evaluating Contextual Informativeness}
Two pieces of prior work also focus on the automatic evaluation of contextual informativeness. The first formalizes the task and introduces a crowd-sourced dataset that uses a Likert scale as a gold standard for contextual informativeness scores \cite{Kapelner_2018}. The second work experiments with this dataset and proposes an attention-based model to create vector representations of both the word and its surrounding context \cite{Nam2022}, which provide the basis for predicting informativeness scores. 
This model achieves strong performance on adult-directed data which includes a single target word in each passage.

\section{Task and Data}
In this section, we describe the creation of our dataset and formally introduce the task of \textit{measuring contextual informativeness in children's stories}.

\subsection{Dataset} \label{ssec:dataset}
Our dataset builds on \citet{valentini}, which contains 180 LLM-generated children's stories. Each story utilizes five target vocabulary words selected based on age of acquisition which the LLMs have been tasked to include. We annotate how much contextual support is provided for each target word. 

Our annotation schema is a modified version of the cloze task, in which annotators fill in blanks with the words they believe best complete a story. As the stories from \citet{valentini} each contain five target words, all target words are replaced with blanks labeled 1 to 5 to simulate a child's incomprehension of the unknown targets. Annotators guess the missing word for each number; there may be one or more blanks for each number. 

The cloze task traditionally only accepts one correct answer and therefore fails to reward relevant alternatives (e.g., synonyms and hypernyms). To address this,
we score based on the semantic similarity between the predicted and actual word. We calculate the cosine similarity between the word embedding of each guess and true target word using ConceptNet Numberbatch 19.08 English embeddings \cite{speer2017conceptnet}.\footnote{For the rationale behind our choice of word embeddings, please refer to Appendix \ref{appendix:wordembeddings}.} This similarity is averaged across guesses from three annotators. The resulting value is intuitively indicative of how well annotators are able to narrow the semantic space of the missing word based on its context.\footnote{Please refer to Appendix \ref{appendix:annotators} for annotation instructions.}

We have six university-level, fluent English speakers annotate all target words for 60 to 180 stories such that each story has three annotators. We manually review annotations and exclude all stories with insufficient or unsatisfactory responses, resulting in a final dataset of 765 target words across 153 generated stories. 

\subsection{Formal Task Definition}
We define the task as \textit{measuring contextual informativeness in children's stories}, focusing on passages with multiple target words, each potentially occurring more than once. \textit{Contextual informativeness} refers to the extent to which the surrounding words and phrases clarify the meaning or usage of a target vocabulary item. 

Given a set of stories $S = \{S_1,S_2,...,S_m\}$ with target vocabulary words $T_i = \{t_{i,1}, t_{i,2}, ..., t_{i,n}\}$ where $i \in [1, m]$, the goal is to evaluate the contextual informativeness of each passage $S_i$ with respect to all instances of a target word $t_{i,j}$. 

The dataset consists of $m*n$ instances represented as $(S_i, T_i, t_{i,j}, c_{i,j})$ where $c_{i,j}$ is the gold standard informativeness score for a target word $t_{i,j}$ in the context of $S_i$. We aim to learn the function $C(S_i,T_i,t_{i,j};\theta)$ that predicts the contextual informativeness of target word $t_{i,j}$ within a story $S_i$ under the constraint that the meanings of the remaining vocabulary terms $T_i/\{t_{i,j}\}$ are unknown at inference time. The task is evaluated primarily on the Spearman's rank correlation coefficient $\rho(\hat{c}, c)$, where higher values indicate stronger agreement between the predicted level of contextual informativeness $\hat{c}_{i,j} = C(S_i,T_i,t_{i,j};\theta)$ and the gold standard score $c_{i,j}$ across all passages $S_i \in S$ and their associated target vocabulary terms $t_{i,j} \in T_i$.

\section{Experiments}

\subsection{Proposed Methods}
In our approaches to the specified task, we leverage the capabilities of RoBERTa and Gemini to simulate extracting contextual information and articulating it as a guess, as performed by our annotators. Predicting masked words mirrors the human ability to extract relevant information and allows us to estimate the contextual informativeness of a text.

Our first approach is based on RoBERTa \cite{roberta}, a masked language model (MLM) we expect to be highly suitable for our task. RoBERTa's architecture is based on a transformer model. We use RoBERTa by predicting words and computing the word embedding similarity of the word embeddings corresponding to the target word and the ground truth. We employ RoBERTa through the transformers library \cite{wolf-etal-2020-transformers}. Full details are shown in Appendix \ref{appendix:computationalexperimentdetails}.

One challenge of our task setup consists of combining  predictions for multiple masked instances of the same word and hiding instances of other "unknown" (i.e., to a child) target words in each passage. To combine multiple word occurrences, we lemmatize the predictions made for each mask infill. We then combine the individual predictions corresponding to the same lemma by summing the probabilities and getting the overall top prediction based on the lemma with the highest cumulative score. To hide instances of unknown words, we replace any additional target words with the unknown token. 
This approach is denoted by \textbf{RoBERTa-mult}.

\begin{table*}
\centering
\begin{tabular}{l|cc|cc|cc|cc|cc}

\midrule

& \multicolumn{2}{c|}{\textbf{Spearman's $\rho$}}                                   
& \multicolumn{2}{c|}{\textbf{$\rho$-significance}}  

& \multicolumn{2}{c|}{\textbf{Pearson's $r$}}  

& \multicolumn{2}{c|}{\textbf{$r$-significance}}

& \multicolumn{2}{c}{\textbf{RMSE}}
\\ 
\midrule
\textbf{Context Similarity}                             
& \multicolumn{2}{c|}{0.2890}                            
& \multicolumn{2}{c|}{$3.68 \times 10^{-16}$}  

& \multicolumn{2}{c|}{0.2858}  

& \multicolumn{2}{c|}{$7.91 \times 10^{-16}$} 

& \multicolumn{2}{c}{0.3078}  
\\

\textbf{Context Window}                                

& \multicolumn{2}{c|}{0.3134}                                         
& \multicolumn{2}{c|}{$7.08 \times 10^{-19}$}  

& \multicolumn{2}{c|}{0.2772}   

& \multicolumn{2}{c|}{$6.06 \times 10^{-15}$} 

& \multicolumn{2}{c}{0.2921} 
\\

\textbf{Num Related Words}                             

& \multicolumn{2}{c|}{0.3534}                                     
& \multicolumn{2}{c|}{$6.82 \times 10^{-24}$}  

& \multicolumn{2}{c|}{0.3239} 

& \multicolumn{2}{c|}{$4.03 \times 10^{-20}$} 

& \multicolumn{2}{c}{0.4120} 
\\
\textbf{Nam et al.}                                

& \multicolumn{2}{c|}{0.0525}                                         
& \multicolumn{2}{c|}{0.1472}      

& \multicolumn{2}{c|}{0.0505}  

& \multicolumn{2}{c|}{0.1635}  

& \multicolumn{2}{c}{0.3165}  
\\

\textbf{Nam et al.+WordNet}                         

& \multicolumn{2}{c|}{0.0574}                                     
& \multicolumn{2}{c|}{0.1127}  

& \multicolumn{2}{c|}{0.0623} 

& \multicolumn{2}{c|}{0.0850}

& \multicolumn{2}{c}{0.3166}
\\
\textbf{RoBERTa-mult}                         
  
& \multicolumn{2}{c|}{0.4601}                                     
& \multicolumn{2}{c|}{$2.72 \times 10^{-41}$}  

& \multicolumn{2}{c|}{0.4721}  

& \multicolumn{2}{c|}{$1.18 \times 10^{-43}$} 

& \multicolumn{2}{c}{0.2972}    

\\
\textbf{Gemini}                         

& \multicolumn{2}{c|}{\textbf{0.4983}}                                       
& \multicolumn{2}{c|}{$3.39 \times 10^{-49}$}

& \multicolumn{2}{c|}{\textbf{0.5297}}  

& \multicolumn{2}{c|}{$1.80 \times 10^{-56}$}

& \multicolumn{2}{c}{\textbf{0.2870}}
\\ \bottomrule
\end{tabular}

\caption{Full results on our dataset, for all models and baselines. $N$-significance refers to the reported p-value for each correlation metric $N$.}
\label{table1}
\end{table*}

Our other proposed approach uses \textbf{Gemini}, a state-of-the-art LLM from Google \cite{Gemini}. We examine the use of an LLM in place of an MLM to see whether it will perform better at the task due to its massive amounts of pretraining data. The model is provided prompts in the following style:

\begin{itemize}
    \item In the following story, guess the word that is replaced by '<mask>'. Ignore any other blanks ($\_\_\_\_$) and ONLY try to guess the word replaced by '<mask>'.
\end{itemize}

For both approaches, the informativeness score is obtained by calculating the ConceptNet similarity between the guessed word and the missing target.

\subsection{Baselines}

We compare our model to various simple baselines and existing models.

\paragraph{Context Similarity}
Context similarity refers to the average cosine similarity of the word embedding of every word in a passage or story and the target word. We exclude stop words, other instances of the target, and instances of any other target words in the text.

\paragraph{Context Window}
We then consider the words only directly surrounding the target in a window of five words on either side of the target. We average the cosine similarity of each word in the window and the target. If a stop word or target appears in the window, the window is adjusted to include the next word in order to retain its size when possible. 

\paragraph{Number of Related Words}
We further consider the number of words that have a cosine similarity with the target above a threshold of 0.3,\footnote{We initially experiment with multiple thresholds as well as context window sizes, including only the best performing in the results. See Appendix \ref{appendix:contextbaselines} for full context baseline results.} excluding the stop words and targets as in the prior baselines. 

\paragraph{Nam et al. Model} We also compare to the model proposed by \citet{Nam2022}, which is trained on the gold standard from \citet{Kapelner_2018} and achieves state-of-the-art results on adult-directed data. Notably, our task differs from theirs in our focus on children's stories. In addition, we use significantly longer contexts and model the occurrence of multiple target words within the same story context. Nonetheless, we test the model on our primary dataset to see if it can generalize to child-directed text.

\paragraph{Modified Nam et al. Model} In addition to the base model provided by \citet{Nam2022}, we also experiment with a slightly modified version which adds information about the target word using its WordNet vector \cite{Saedi2018}. 
We expect this to improve the model as the base model does not obtain any information about the target word. This approach is denoted by \textbf{Nam et al.+WordNet}.

\subsection{Metrics}
We evaluate all models and baselines using the following three metrics: Pearson's $r$, Spearman's $\rho$, and root-mean-square error (RMSE). We consider Spearman's $\rho$ to be our main metric, as it assesses monotonic relationships that can be linear or nonlinear, where Pearson's $r$ 
measures linear relationships. We do not consider RMSE to be a primary metric for this task as it loses comparative power in edge cases and with discrete variables, but we include it as it is the only metric reported by \citet{Nam2022}.

\subsection{Results}
Our results, shown in full in Table \ref{table1}, demonstrate that 
using Gemini is the best performing method for the task: 
it achieves a Spearman correlation coefficient of 0.4983 with our gold standard annotations, a Pearson correlation coefficient of 0.5297, and an RMSE of 0.2870.

\begin{table*}
\centering
\begin{tabular}{l|cc|cc|cc|cc|cc}
\midrule

& \multicolumn{2}{c|}{\textbf{Spearman's $\rho$}}                                   
& \multicolumn{2}{c|}{\textbf{$\rho$-significance}}   

& \multicolumn{2}{c|}{\textbf{Pearson's $r$}}  

& \multicolumn{2}{c|}{\textbf{$r$-significance}}

& \multicolumn{2}{c}{\textbf{RMSE}}                                                  
\\ 
\midrule
\textbf{Context Similarity}                            

& \multicolumn{2}{c|}{0.2287}                                     
& \multicolumn{2}{c|}{0.0011}

& \multicolumn{2}{c|}{0.2314}

& \multicolumn{2}{c|}{0.0010} 

& \multicolumn{2}{c}{0.2722} 

\\ 
\textbf{Context Window}                                

& \multicolumn{2}{c|}{0.2345}

& \multicolumn{2}{c|}{0.0008}

& \multicolumn{2}{c|}{0.2778}

& \multicolumn{2}{c|}{$6.82 \times 10^{-5}$} 

& \multicolumn{2}{c}{\textbf{0.2573}}                                       
\\

\textbf{Num Related Words}                             

& \multicolumn{2}{c|}{0.2797}

& \multicolumn{2}{c|}{$6.06 \times 10^{-5}$}

& \multicolumn{2}{c|}{0.2583}

& \multicolumn{2}{c|}{0.0002}

& \multicolumn{2}{c}{0.3466}                                         
\\
\textbf{Nam et al.}                                

& \multicolumn{2}{c|}{0.3545}                                     
& \multicolumn{2}{c|}{$2.61 \times 10^{-7}$}

& \multicolumn{2}{c|}{0.3217}                                     
& \multicolumn{2}{c|}{$3.40 \times 10^{-6}$}

& \multicolumn{2}{c}{0.3971}                                         
\\

\textbf{Nam et al.+WordNet}                         

& \multicolumn{2}{c|}{0.3660}                                     
& \multicolumn{2}{c|}{$9.87 \times 10^{-8}$}

& \multicolumn{2}{c|}{0.3230}     

& \multicolumn{2}{c|}{$3.09 \times 10^{-6}$}

& \multicolumn{2}{c}{0.3540}                                                                                                         
\\

\textbf{RoBERTa-mult}                         

& \multicolumn{2}{c|}{0.3796}                                     
& \multicolumn{2}{c|}{$2.97 \times 10^{-8}$}

& \multicolumn{2}{c|}{0.3886}                                     
& \multicolumn{2}{c|}{$1.30 \times 10^{-8}$}

& \multicolumn{2}{c}{0.2715}         

\\
\textbf{Gemini}                         

& \multicolumn{2}{c|}{\textbf{0.3908}}                              
& \multicolumn{2}{c|}{$1.05 \times 10^{-8}$}

& \multicolumn{2}{c|}{\textbf{0.4209}}                

& \multicolumn{2}{c|}{$5.42 \times 10^{-10}$}

& \multicolumn{2}{c}{0.3651} 

\\ \bottomrule
\end{tabular}
\caption{Full results on the Kapelner dataset, for all models and baselines. $N$-significance refers to the reported p-value for each correlation metric $N$.}
\label{table2}
\end{table*}

RoBERTa-mult performs only slightly worse than Gemini, with a Spearman correlation of 0.4601, a Pearson correlation coefficient of 0.4721, and an RMSE of 0.2972. 

We find that, on our dataset, \citet{Nam2022} underperforms the simple baselines. This is reasonable, as it is a model trained on adult-directed text. We expect it to have relatively poor generalization abilities given that it is an attention-based model trained on a specific dataset.

\section{Analysis: Generalization Abilities}
We further aim to see if our proposed approaches generalize to measuring contextual informativeness in adult-directed text. 

\paragraph{Dataset} 
We leverage the dataset from \citet{Kapelner_2018} with adult-directed text instances (and corresponding target words) for contextual informativeness evaluation. Importantly, the target words are more complex than in our dataset of children's stories, and it generally contains more advanced language. The original annotations differ from ours (see \citet{Kapelner_2018} for a description), 
so, for comparability reasons, we re-annotate 200 contexts from that dataset using our annotation schema. Each instance is annotated by two independent annotators, and the similarity scores are averaged.

\paragraph{Results} 

Results from our analysis (shown in Table \ref{table2}) demonstrate that both proposed methods generalize well to adult-directed text: they achieve a moderate correlation with the ground truth on the re-annotated portion of the Kapelner dataset and outperform all baselines and models for both correlation metrics. For RMSE, \textit{context window} achieves the strongest score, closely followed by RoBERTa-mult. This suggests that, while Gemini (RMSE = 0.3651) effectively identifies which passages are more or less contextually informative, it struggles with calculating exact values for this dataset.

These results indicate that our previous finding (that the best performing methods initially reported on the adult-directed data do not generalize well to child-directed data) does not hold true in the converse. The attention-based model achieves the best scores on the original Likert scale-based Kapelner annotations,\footnote{Full results of \citet{Nam2022}, including on the Kapelner dataset using their Likert scale-based annotation schema are located in Appendix \ref{appendix:likert}.} and the correlation is only slightly lower than that of Gemini and RoBERTa-mult when using our annotation schema. However, on the child-directed dataset, the Spearman coefficient drops to only 0.0574, exhibiting almost no correlation at all. 

\section{Conclusion}

We propose the task of \textit{measuring contextual informativeness in children's stories} with respect to target vocabulary words. We provide a task definition, along with a gold standard dataset for the task. As methods to address the task, we test RoBERTa and Gemini by using the similarity of their predictions to the true target words to produce a contextual informativeness score.
On the child-directed dataset, Gemini achieves a Spearman's rank correlation of 0.4983, while the highest performing baseline only obtains 0.3534. We further show that our method generalizes well to adult-directed text, once again outperforming all baselines.

These findings highlight the potential of automated methods for evaluating and improving the educational value of children's stories. We hope this work serves as a strong starting point for future research on the automatic assessment and optimization of vocabulary learning tools, particularly as needed for the automatic generation of personalized vocabulary intervention materials in early childhood.

\section*{Limitations}

Because the creation of our dataset and modification of the Kapelner dataset required human annotators, all of which were undergraduate or graduate student volunteers, something that could greatly strengthen this work in the future would be the use of additional annotators to add more statistical significance to our results.

The use of additional models or additional methods for evaluating these models (e.g., perplexity), could also yield more insights and is encouraged as a direction for future work. 

Finally, while contextual informativeness with respect to target words is important to measure, it does not necessarily correlate with the learnability of those words for children who read the stories. In future work, we hope to use data from ongoing experiments to bridge the gap between contextual informativeness and vocabulary learnability in early childhood.

\section*{Ethics Statement}

No data involved uses any sort of personal information and is all either available to the public or used with full permission and knowledge of intended use from the authors. 

In terms of other ethical considerations, we find that the risks of this study are minimal to none. Though the results of this research are eventually intended for children, no vulnerable populations were involved in this study up to this point. If automatically generated stories are given or read to children, it is important to verify in advance that they are safe for the target population, as current models cannot guarantee this.

\section*{Acknowledgments}
We thank the anonymous reviewers for their helpful comments. This research was supported by the NSF under grant IIS 2223917. The opinions expressed are those of the authors and do not represent views of the NSF.

\bibliography{custom}

\begin{thebibliography}{25}
\providecommand{\natexlab}[1]{#1}

\bibitem[{Brysbaert et~al.(2016)Brysbaert, Stevens, Mandera, and Keuleers}]{Brysbaert_Stevens_Mandera_Keuleers_2016}
Marc Brysbaert, Michaël Stevens, Paweł Mandera, and Emmanuel Keuleers. 2016.
\newblock \href {https://www.frontiersin.org/articles/10.3389/fpsyg.2016.01116} {How many words do we know? practical estimates of vocabulary size dependent on word definition, the degree of language input and the participant’s age}.
\newblock \emph{Frontiers in Psychology}, 7.

\bibitem[{Carlisle and Rice(2004)}]{carlisle2004assessment}
J~Carlisle and M~Rice. 2004.
\newblock Assessment of reading comprehension.
\newblock \emph{Handbook of language and literacy}, pages 521--555.

\bibitem[{Duff et~al.(2015)Duff, Tomblin, and Catts}]{Duff_Tomblin_Catts_2015}
Dawna Duff, J.~Bruce Tomblin, and Hugh Catts. 2015.
\newblock \href {https://doi.org/10.1044/2015_JSLHR-L-13-0310} {The influence of reading on vocabulary growth: A case for a matthew effect}.
\newblock \emph{Journal of Speech, Language, and Hearing Research: JSLHR}, 58(3):853–864.

\bibitem[{Finkelstein et~al.(2001)Finkelstein, Gabrilovich, Matias, Rivlin, Solan, Wolfman, and Ruppin}]{finkelstein2001placing}
Lev Finkelstein, Evgeniy Gabrilovich, Yossi Matias, Ehud Rivlin, Zach Solan, Gadi Wolfman, and Eytan Ruppin. 2001.
\newblock Placing search in context: The concept revisited.
\newblock In \emph{Proceedings of the 10th international conference on World Wide Web}, pages 406--414.

\bibitem[{Gellert and Elbro(2013)}]{Gellert_Elbro_2013}
Anna~S. Gellert and Carsten Elbro. 2013.
\newblock \href {https://doi.org/10.1177/0734282912451971} {Cloze tests may be quick, but are they dirty? development and preliminary validation of a cloze test of reading comprehension}.
\newblock \emph{Journal of Psychoeducational Assessment}, 31(1):16–28.

\bibitem[{Gemini-Team et~al.(2024)Gemini-Team, Anil, Borgeaud, Alayrac, Yu, Soricut, Schalkwyk, Dai, Hauth, Millican, Silver, Johnson, Antonoglou, Schrittwieser, Glaese, Chen, Pitler, Lillicrap, Lazaridou, Firat, Molloy, Isard, Barham, Hennigan, Lee, Viola, Reynolds, Xu, Doherty, Collins, Meyer, Rutherford, Moreira, Ayoub, Goel, Krawczyk, Du, Chi, Cheng, Ni, Shah, Kane, Chan, Faruqui, Severyn, Lin, Li, Cheng, Ittycheriah, Mahdieh, Chen, Sun, Tran, Bagri, Lakshminarayanan, Liu, Orban, Güra, Zhou, Song, Boffy, Ganapathy, Zheng, Choe, Ágoston Weisz, Zhu, Lu, Gopal, Kahn, Kula, Pitman, Shah, Taropa, Merey, Baeuml, Chen, Shafey, Zhang, Sercinoglu, Tucker, Piqueras, Krikun, Barr, Savinov, Danihelka, Roelofs, White, Andreassen, von Glehn, Yagati, Kazemi, Gonzalez, Khalman, Sygnowski, Frechette, Smith, Culp, Proleev, Luan, Chen, Lottes, Schucher, Lebron, Rrustemi, Clay, Crone, Kocisky, Zhao, Perz, Yu, Howard, Bloniarz, Rae, Lu, Sifre, Maggioni, Alcober, Garrette, Barnes, Thakoor, Austin, Barth-Maron, Wong, Joshi,
  Chaabouni, Fatiha, Ahuja, Tomar, Senter, Chadwick, Kornakov, Attaluri, Iturrate, Liu, Li, Cogan, Chen, Jia, Gu, Zhang, Grimstad, Hartman, Garcia, Pillai, Devlin, Laskin, de~Las~Casas, Valter, Tao, Blanco, Badia, Reitter, Chen, Brennan, Rivera, Brin, Iqbal, Surita, Labanowski, Rao, Winkler, Parisotto, Gu, Olszewska, Addanki, Miech, Louis, Teplyashin, Brown, Catt, Balaguer, Xiang, Wang, Ashwood, Briukhov, Webson, Ganapathy, Sanghavi, Kannan, Chang, Stjerngren, Djolonga, Sun, Bapna, Aitchison, Pejman, Michalewski, Yu, Wang, Love, Ahn, Bloxwich, Han, Humphreys, Sellam, Bradbury, Godbole, Samangooei, Damoc, Kaskasoli, Arnold, Vasudevan, Agrawal, Riesa, Lepikhin, Tanburn, Srinivasan, Lim, Hodkinson, Shyam, Ferret, Hand, Garg, Paine, Li, Li, Giang, Neitz, Abbas, York, Reid, Cole, Chowdhery, Das, Rogozińska, Nikolaev, Sprechmann, Nado, Zilka, Prost, He, Monteiro, Mishra, Welty, Newlan, Jia, Allamanis, Hu, de~Liedekerke, Gilmer, Saroufim, Rijhwani, Hou, Shrivastava, Baddepudi, Goldin, Ozturel, Cassirer, Xu, Sohn,
  Sachan, Amplayo, Swanson, Petrova, Narayan, Guez, Brahma, Landon, Patel, Zhao, Villela, Wang, Jia, Rahtz, Giménez, Yeung, Keeling, Georgiev, Mincu, Wu, Haykal, Saputro, Vodrahalli, Qin, Cankara, Sharma, Fernando, Hawkins, Neyshabur, Kim, Hutter, Agrawal, Castro-Ros, van~den Driessche, Wang, Yang, yiin Chang, Komarek, McIlroy, Lučić, Zhang, Farhan, Sharman, Natsev, Michel, Bansal, Qiao, Cao, Shakeri, Butterfield, Chung, Rubenstein, Agrawal, Mensch, Soparkar, Lenc, Chung, Pope, Maggiore, Kay, Jhakra, Wang, Maynez, Phuong, Tobin, Tacchetti, Trebacz, Robinson, Katariya, Riedel, Bailey, Xiao, Ghelani, Aroyo, Slone, Houlsby, Xiong, Yang, Gribovskaya, Adler, Wirth, Lee, Li, Kagohara, Pavagadhi, Bridgers, Bortsova, Ghemawat, Ahmed, Liu, Powell, Bolina, Iinuma, Zablotskaia, Besley, Chung, Dozat, Comanescu, Si, Greer, Su, Polacek, Kaufman, Tokumine, Hu, Buchatskaya, Miao, Elhawaty, Siddhant, Tomasev, Xing, Greer, Miller, Ashraf, Roy, Zhang, Ma, Filos, Besta, Blevins, Klimenko, Yeh, Changpinyo, Mu, Chang,
  Pajarskas, Muir, Cohen, Lan, Haridasan, Marathe, Hansen, Douglas, Samuel, Wang, Austin, Lan, Jiang, Chiu, Lorenzo, Sjösund, Cevey, Gleicher, Avrahami, Boral, Srinivasan, Selo, May, Aisopos, Hussenot, Soares, Baumli, Chang, Recasens, Caine, Pritzel, Pavetic, Pardo, Gergely, Frye, Ramasesh, Horgan, Badola, Kassner, Roy, Dyer, Campos, Tomala, Tang, Badawy, White, Mustafa, Lang, Jindal, Vikram, Gong, Caelles, Hemsley, Thornton, Feng, Stokowiec, Zheng, Thacker, Çağlar Ünlü, Zhang, Saleh, Svensson, Bileschi, Patil, Anand, Ring, Tsihlas, Vezer, Selvi, Shevlane, Rodriguez, Kwiatkowski, Daruki, Rong, Dafoe, FitzGerald, Gu-Lemberg, Khan, Hendricks, Pellat, Feinberg, Cobon-Kerr, Sainath, Rauh, Hashemi, Ives, Hasson, Noland, Cao, Byrd, Hou, Wang, Sottiaux, Paganini, Lespiau, Moufarek, Hassan, Shivakumar, van Amersfoort, Mandhane, Joshi, Goyal, Tung, Brock, Sheahan, Misra, Li, Rakićević, Dehghani, Liu, Mittal, Oh, Noury, Sezener, Huot, Lamm, Cao, Chen, Mudgal, Stella, Brooks, Vasudevan, Liu, Chain, Melinkeri,
  Cohen, Wang, Seymore, Zubkov, Goel, Yue, Krishnakumaran, Albert, Hurley, Sano, Mohananey, Joughin, Filonov, Kępa, Eldawy, Lim, Rishi, Badiezadegan, Bos, Chang, Jain, Padmanabhan, Puttagunta, Krishna, Baker, Kalb, Bedapudi, Kurzrok, Lei, Yu, Litvin, Zhou, Wu, Sobell, Siciliano, Papir, Neale, Bragagnolo, Toor, Chen, Anklin, Wang, Feng, Gholami, Ling, Liu, Walter, Moghaddam, Kishore, Adamek, Mercado, Mallinson, Wandekar, Cagle, Ofek, Garrido, Lombriser, Mukha, Sun, Mohammad, Matak, Qian, Peswani, Janus, Yuan, Schelin, David, Garg, He, Duzhyi, Älgmyr, Lottaz, Li, Yadav, Xu, Chinien, Shivanna, Chuklin, Li, Spadine, Wolfe, Mohamed, Das, Dai, He, von Dincklage, Upadhyay, Maurya, Chi, Krause, Salama, Rabinovitch, M, Selvan, Dektiarev, Ghiasi, Guven, Gupta, Liu, Sharma, Shtacher, Paul, Akerlund, Aubet, Huang, Zhu, Zhu, Teixeira, Fritze, Bertolini, Marinescu, Bölle, Paulus, Gupta, Latkar, Chang, Sanders, Wilson, Wu, Tan, Thiet, Doshi, Lall, Mishra, Chen, Luong, Benjamin, Lee, Andrejczuk, Rabiej, Ranjan, Styrc,
  Yin, Simon, Harriott, Bansal, Robsky, Bacon, Greene, Mirylenka, Zhou, Sarvana, Goyal, Andermatt, Siegler, Horn, Israel, Pongetti, Chen, Selvatici, Silva, Wang, Tolins, Guu, Yogev, Cai, Agostini, Shah, Nguyen, Donnaile, Pereira, Friso, Stambler, Kurzrok, Kuang, Romanikhin, Geller, Yan, Jang, Lee, Fica, Malmi, Tan, Banica, Balle, Pham, Huang, Avram, Shi, Singh, Hidey, Ahuja, Saxena, Dooley, Potharaju, O'Neill, Gokulchandran, Foley, Zhao, Dusenberry, Liu, Mehta, Kotikalapudi, Safranek-Shrader, Goodman, Kessinger, Globen, Kolhar, Gorgolewski, Ibrahim, Song, Eichenbaum, Brovelli, Potluri, Lahoti, Baetu, Ghorbani, Chen, Crawford, Pal, Sridhar, Gurita, Mujika, Petrovski, Cedoz, Li, Chen, Santo, Goyal, Punjabi, Kappaganthu, Kwak, LV, Velury, Choudhury, Hall, Shah, Figueira, Thomas, Lu, Zhou, Kumar, Jurdi, Chikkerur, Ma, Yu, Kwak, Ähdel, Rajayogam, Choma, Liu, Barua, Ji, Park, Hellendoorn, Bailey, Bilal, Zhou, Khatir, Sutton, Rzadkowski, Macintosh, Shagin, Medina, Liang, Zhou, Shah, Bi, Dankovics, Banga, Lehmann,
  Bredesen, Lin, Hoffmann, Lai, Chung, Yang, Balani, Bražinskas, Sozanschi, Hayes, Alcalde, Makarov, Chen, Stella, Snijders, Mandl, Kärrman, Nowak, Wu, Dyck, Vaidyanathan, R, Mallet, Rudominer, Johnston, Mittal, Udathu, Christensen, Verma, Irving, Santucci, Elsayed, Davoodi, Georgiev, Tenney, Hua, Cideron, Leurent, Alnahlawi, Georgescu, Wei, Zheng, Scandinaro, Jiang, Snoek, Sundararajan, Wang, Ontiveros, Karo, Cole, Rajashekhar, Tumeh, Ben-David, Jain, Uesato, Datta, Bunyan, Wu, Zhang, Stanczyk, Zhang, Steiner, Naskar, Azzam, Johnson, Paszke, Chiu, Elias, Mohiuddin, Muhammad, Miao, Lee, Vieillard, Park, Zhang, Stanway, Garmon, Karmarkar, Dong, Lee, Kumar, Zhou, Evens, Isaac, Irving, Loper, Fink, Arkatkar, Chen, Shafran, Petrychenko, Chen, Jia, Levskaya, Zhu, Grabowski, Mao, Magni, Yao, Snaider, Casagrande, Palmer, Suganthan, Castaño, Giannoumis, Kim, Rybiński, Sreevatsa, Prendki, Soergel, Goedeckemeyer, Gierke, Jafari, Gaba, Wiesner, Wright, Wei, Vashisht, Kulizhskaya, Hoover, Le, Li, Iwuanyanwu, Liu,
  Ramirez, Khorlin, Cui, LIN, Wu, Aguilar, Pallo, Chakladar, Perng, Abellan, Zhang, Dasgupta, Kushman, Penchev, Repina, Wu, van~der Weide, Ponnapalli, Kaplan, Simsa, Li, Dousse, Yang, Piper, Ie, Pasumarthi, Lintz, Vijayakumar, Andor, Valenzuela, Lui, Paduraru, Peng, Lee, Zhang, Greene, Nguyen, Kurylowicz, Hardin, Dixon, Janzer, Choo, Feng, Zhang, Singhal, Du, McKinnon, Antropova, Bolukbasi, Keller, Reid, Finchelstein, Raad, Crocker, Hawkins, Dadashi, Gaffney, Franko, Bulanova, Leblond, Chung, Askham, Cobo, Xu, Fischer, Xu, Sorokin, Alberti, Lin, Evans, Dimitriev, Forbes, Banarse, Tung, Omernick, Bishop, Sterneck, Jain, Xia, Amid, Piccinno, Wang, Banzal, Mankowitz, Polozov, Krakovna, Brown, Bateni, Duan, Firoiu, Thotakuri, Natan, Geist, tan Girgin, Li, Ye, Roval, Tojo, Kwong, Lee-Thorp, Yew, Sinopalnikov, Ramos, Mellor, Sharma, Wu, Miller, Sonnerat, Vnukov, Greig, Beattie, Caveness, Bai, Eisenschlos, Korchemniy, Tsai, Jasarevic, Kong, Dao, Zheng, Liu, Yang, Zhu, Teh, Sanmiya, Gladchenko, Trdin, Toyama, Rosen,
  Tavakkol, Xue, Elkind, Woodman, Carpenter, Papamakarios, Kemp, Kafle, Grunina, Sinha, Talbert, Wu, Owusu-Afriyie, Du, Thornton, Pont-Tuset, Narayana, Li, Fatehi, Wieting, Ajmeri, Uria, Ko, Knight, Héliou, Niu, Gu, Pang, Li, Levine, Stolovich, Santamaria-Fernandez, Goenka, Yustalim, Strudel, Elqursh, Deck, Lee, Li, Levin, Hoffmann, Holtmann-Rice, Bachem, Arora, Koh, Yeganeh, Põder, Tariq, Sun, Ionita, Seyedhosseini, Tafti, Liu, Gulati, Liu, Ye, Chrzaszcz, Wang, Sethi, Li, Brown, Singh, Fan, Parisi, Stanton, Koverkathu, Choquette-Choo, Li, Lu, Ittycheriah, Shroff, Varadarajan, Bahargam, Willoughby, Gaddy, Desjardins, Cornero, Robenek, Mittal, Albrecht, Shenoy, Moiseev, Jacobsson, Ghaffarkhah, Rivière, Walton, Crepy, Parrish, Zhou, Farabet, Radebaugh, Srinivasan, van~der Salm, Fidjeland, Scellato, Latorre-Chimoto, Klimczak-Plucińska, Bridson, de~Cesare, Hudson, Mendolicchio, Walker, Morris, Mauger, Guseynov, Reid, Odoom, Loher, Cotruta, Yenugula, Grewe, Petrushkina, Duerig, Sanchez, Yadlowsky, Shen,
  Globerson, Webb, Dua, Li, Bhupatiraju, Hurt, Qureshi, Agarwal, Shani, Eyal, Khare, Belle, Wang, Tekur, Kale, Wei, Sang, Saeta, Liechty, Sun, Zhao, Lee, Nayak, Fritz, Vuyyuru, Aslanides, Vyas, Wicke, Ma, Eltyshev, Martin, Cate, Manyika, Amiri, Kim, Xiong, Kang, Luisier, Tripuraneni, Madras, Guo, Waters, Wang, Ainslie, Baldridge, Zhang, Pruthi, Bauer, Yang, Mansour, Gelman, Xu, Polovets, Liu, Cai, Chen, Sheng, Xue, Ozair, Angermueller, Li, Sinha, Wang, Wiesinger, Koukoumidis, Tian, Iyer, Gurumurthy, Goldenson, Shah, Blake, Yu, Urbanowicz, Palomaki, Fernando, Durden, Mehta, Momchev, Rahimtoroghi, Georgaki, Raul, Ruder, Redshaw, Lee, Zhou, Jalan, Li, Hechtman, Schuh, Nasr, Milan, Mikulik, Franco, Green, Nguyen, Kelley, Mahendru, Hu, Howland, Vargas, Hui, Bansal, Rao, Ghiya, Wang, Ye, Sarr, Preston, Elish, Li, Kaku, Gupta, Pasupat, Juan, Someswar, M., Chen, Amini, Fabrikant, Chu, Dong, Muthal, Buthpitiya, Jauhari, Hua, Khandelwal, Hitron, Ren, Rinaldi, Drath, Dabush, Jiang, Godhia, Sachs, Chen, Fan, Taitelbaum,
  Noga, Dai, Wang, Liang, Hamer, Ferng, Elkind, Atias, Lee, Listík, Carlen, van~de Kerkhof, Pikus, Zaher, Müller, Zykova, Stefanec, Gatsko, Hirnschall, Sethi, Xu, Ahuja, Tsai, Stefanoiu, Feng, Dhandhania, Katyal, Gupta, Parulekar, Pitta, Zhao, Bhatia, Bhavnani, Alhadlaq, Li, Danenberg, Tu, Pine, Filippova, Ghosh, Limonchik, Urala, Lanka, Clive, Sun, Li, Wu, Hongtongsak, Li, Thakkar, Omarov, Majmundar, Alverson, Kucharski, Patel, Jain, Zabelin, Pelagatti, Kohli, Kumar, Kim, Sankar, Shah, Ramachandruni, Zeng, Bariach, Weidinger, Vu, Andreev, He, Hui, Kashem, Subramanya, Hsiao, Hassabis, Kavukcuoglu, Sadovsky, Le, Strohman, Wu, Petrov, Dean, and Vinyals}]{Gemini}
Gemini-Team, Rohan Anil, Sebastian Borgeaud, Jean-Baptiste Alayrac, Jiahui Yu, Radu Soricut, Johan Schalkwyk, Andrew~M. Dai, Anja Hauth, Katie Millican, David Silver, Melvin Johnson, Ioannis Antonoglou, Julian Schrittwieser, Amelia Glaese, Jilin Chen, Emily Pitler, Timothy Lillicrap, Angeliki Lazaridou, Orhan Firat, James Molloy, Michael Isard, Paul~R. Barham, Tom Hennigan, Benjamin Lee, Fabio Viola, Malcolm Reynolds, Yuanzhong Xu, Ryan Doherty, Eli Collins, Clemens Meyer, Eliza Rutherford, Erica Moreira, Kareem Ayoub, Megha Goel, Jack Krawczyk, Cosmo Du, Ed~Chi, Heng-Tze Cheng, Eric Ni, Purvi Shah, Patrick Kane, Betty Chan, Manaal Faruqui, Aliaksei Severyn, Hanzhao Lin, YaGuang Li, Yong Cheng, Abe Ittycheriah, Mahdis Mahdieh, Mia Chen, Pei Sun, Dustin Tran, Sumit Bagri, Balaji Lakshminarayanan, Jeremiah Liu, Andras Orban, Fabian Güra, Hao Zhou, Xinying Song, Aurelien Boffy, Harish Ganapathy, Steven Zheng, HyunJeong Choe, Ágoston Weisz, Tao Zhu, Yifeng Lu, Siddharth Gopal, Jarrod Kahn, Maciej Kula, Jeff
  Pitman, Rushin Shah, Emanuel Taropa, Majd~Al Merey, Martin Baeuml, Zhifeng Chen, Laurent~El Shafey, Yujing Zhang, Olcan Sercinoglu, George Tucker, Enrique Piqueras, Maxim Krikun, Iain Barr, Nikolay Savinov, Ivo Danihelka, Becca Roelofs, Anaïs White, Anders Andreassen, Tamara von Glehn, Lakshman Yagati, Mehran Kazemi, Lucas Gonzalez, Misha Khalman, Jakub Sygnowski, Alexandre Frechette, Charlotte Smith, Laura Culp, Lev Proleev, Yi~Luan, Xi~Chen, James Lottes, Nathan Schucher, Federico Lebron, Alban Rrustemi, Natalie Clay, Phil Crone, Tomas Kocisky, Jeffrey Zhao, Bartek Perz, Dian Yu, Heidi Howard, Adam Bloniarz, Jack~W. Rae, Han Lu, Laurent Sifre, Marcello Maggioni, Fred Alcober, Dan Garrette, Megan Barnes, Shantanu Thakoor, Jacob Austin, Gabriel Barth-Maron, William Wong, Rishabh Joshi, Rahma Chaabouni, Deeni Fatiha, Arun Ahuja, Gaurav~Singh Tomar, Evan Senter, Martin Chadwick, Ilya Kornakov, Nithya Attaluri, Iñaki Iturrate, Ruibo Liu, Yunxuan Li, Sarah Cogan, Jeremy Chen, Chao Jia, Chenjie Gu, Qiao Zhang,
  Jordan Grimstad, Ale~Jakse Hartman, Xavier Garcia, Thanumalayan~Sankaranarayana Pillai, Jacob Devlin, Michael Laskin, Diego de~Las~Casas, Dasha Valter, Connie Tao, Lorenzo Blanco, Adrià~Puigdomènech Badia, David Reitter, Mianna Chen, Jenny Brennan, Clara Rivera, Sergey Brin, Shariq Iqbal, Gabriela Surita, Jane Labanowski, Abhi Rao, Stephanie Winkler, Emilio Parisotto, Yiming Gu, Kate Olszewska, Ravi Addanki, Antoine Miech, Annie Louis, Denis Teplyashin, Geoff Brown, Elliot Catt, Jan Balaguer, Jackie Xiang, Pidong Wang, Zoe Ashwood, Anton Briukhov, Albert Webson, Sanjay Ganapathy, Smit Sanghavi, Ajay Kannan, Ming-Wei Chang, Axel Stjerngren, Josip Djolonga, Yuting Sun, Ankur Bapna, Matthew Aitchison, Pedram Pejman, Henryk Michalewski, Tianhe Yu, Cindy Wang, Juliette Love, Junwhan Ahn, Dawn Bloxwich, Kehang Han, Peter Humphreys, Thibault Sellam, James Bradbury, Varun Godbole, Sina Samangooei, Bogdan Damoc, Alex Kaskasoli, Sébastien M.~R. Arnold, Vijay Vasudevan, Shubham Agrawal, Jason Riesa, Dmitry
  Lepikhin, Richard Tanburn, Srivatsan Srinivasan, Hyeontaek Lim, Sarah Hodkinson, Pranav Shyam, Johan Ferret, Steven Hand, Ankush Garg, Tom~Le Paine, Jian Li, Yujia Li, Minh Giang, Alexander Neitz, Zaheer Abbas, Sarah York, Machel Reid, Elizabeth Cole, Aakanksha Chowdhery, Dipanjan Das, Dominika Rogozińska, Vitaliy Nikolaev, Pablo Sprechmann, Zachary Nado, Lukas Zilka, Flavien Prost, Luheng He, Marianne Monteiro, Gaurav Mishra, Chris Welty, Josh Newlan, Dawei Jia, Miltiadis Allamanis, Clara~Huiyi Hu, Raoul de~Liedekerke, Justin Gilmer, Carl Saroufim, Shruti Rijhwani, Shaobo Hou, Disha Shrivastava, Anirudh Baddepudi, Alex Goldin, Adnan Ozturel, Albin Cassirer, Yunhan Xu, Daniel Sohn, Devendra Sachan, Reinald~Kim Amplayo, Craig Swanson, Dessie Petrova, Shashi Narayan, Arthur Guez, Siddhartha Brahma, Jessica Landon, Miteyan Patel, Ruizhe Zhao, Kevin Villela, Luyu Wang, Wenhao Jia, Matthew Rahtz, Mai Giménez, Legg Yeung, James Keeling, Petko Georgiev, Diana Mincu, Boxi Wu, Salem Haykal, Rachel Saputro, Kiran
  Vodrahalli, James Qin, Zeynep Cankara, Abhanshu Sharma, Nick Fernando, Will Hawkins, Behnam Neyshabur, Solomon Kim, Adrian Hutter, Priyanka Agrawal, Alex Castro-Ros, George van~den Driessche, Tao Wang, Fan Yang, Shuo yiin Chang, Paul Komarek, Ross McIlroy, Mario Lučić, Guodong Zhang, Wael Farhan, Michael Sharman, Paul Natsev, Paul Michel, Yamini Bansal, Siyuan Qiao, Kris Cao, Siamak Shakeri, Christina Butterfield, Justin Chung, Paul~Kishan Rubenstein, Shivani Agrawal, Arthur Mensch, Kedar Soparkar, Karel Lenc, Timothy Chung, Aedan Pope, Loren Maggiore, Jackie Kay, Priya Jhakra, Shibo Wang, Joshua Maynez, Mary Phuong, Taylor Tobin, Andrea Tacchetti, Maja Trebacz, Kevin Robinson, Yash Katariya, Sebastian Riedel, Paige Bailey, Kefan Xiao, Nimesh Ghelani, Lora Aroyo, Ambrose Slone, Neil Houlsby, Xuehan Xiong, Zhen Yang, Elena Gribovskaya, Jonas Adler, Mateo Wirth, Lisa Lee, Music Li, Thais Kagohara, Jay Pavagadhi, Sophie Bridgers, Anna Bortsova, Sanjay Ghemawat, Zafarali Ahmed, Tianqi Liu, Richard Powell,
  Vijay Bolina, Mariko Iinuma, Polina Zablotskaia, James Besley, Da-Woon Chung, Timothy Dozat, Ramona Comanescu, Xiance Si, Jeremy Greer, Guolong Su, Martin Polacek, Raphaël~Lopez Kaufman, Simon Tokumine, Hexiang Hu, Elena Buchatskaya, Yingjie Miao, Mohamed Elhawaty, Aditya Siddhant, Nenad Tomasev, Jinwei Xing, Christina Greer, Helen Miller, Shereen Ashraf, Aurko Roy, Zizhao Zhang, Ada Ma, Angelos Filos, Milos Besta, Rory Blevins, Ted Klimenko, Chih-Kuan Yeh, Soravit Changpinyo, Jiaqi Mu, Oscar Chang, Mantas Pajarskas, Carrie Muir, Vered Cohen, Charline~Le Lan, Krishna Haridasan, Amit Marathe, Steven Hansen, Sholto Douglas, Rajkumar Samuel, Mingqiu Wang, Sophia Austin, Chang Lan, Jiepu Jiang, Justin Chiu, Jaime~Alonso Lorenzo, Lars~Lowe Sjösund, Sébastien Cevey, Zach Gleicher, Thi Avrahami, Anudhyan Boral, Hansa Srinivasan, Vittorio Selo, Rhys May, Konstantinos Aisopos, Léonard Hussenot, Livio~Baldini Soares, Kate Baumli, Michael~B. Chang, Adrià Recasens, Ben Caine, Alexander Pritzel, Filip Pavetic,
  Fabio Pardo, Anita Gergely, Justin Frye, Vinay Ramasesh, Dan Horgan, Kartikeya Badola, Nora Kassner, Subhrajit Roy, Ethan Dyer, Víctor~Campos Campos, Alex Tomala, Yunhao Tang, Dalia~El Badawy, Elspeth White, Basil Mustafa, Oran Lang, Abhishek Jindal, Sharad Vikram, Zhitao Gong, Sergi Caelles, Ross Hemsley, Gregory Thornton, Fangxiaoyu Feng, Wojciech Stokowiec, Ce~Zheng, Phoebe Thacker, Çağlar Ünlü, Zhishuai Zhang, Mohammad Saleh, James Svensson, Max Bileschi, Piyush Patil, Ankesh Anand, Roman Ring, Katerina Tsihlas, Arpi Vezer, Marco Selvi, Toby Shevlane, Mikel Rodriguez, Tom Kwiatkowski, Samira Daruki, Keran Rong, Allan Dafoe, Nicholas FitzGerald, Keren Gu-Lemberg, Mina Khan, Lisa~Anne Hendricks, Marie Pellat, Vladimir Feinberg, James Cobon-Kerr, Tara Sainath, Maribeth Rauh, Sayed~Hadi Hashemi, Richard Ives, Yana Hasson, Eric Noland, Yuan Cao, Nathan Byrd, Le~Hou, Qingze Wang, Thibault Sottiaux, Michela Paganini, Jean-Baptiste Lespiau, Alexandre Moufarek, Samer Hassan, Kaushik Shivakumar, Joost van
  Amersfoort, Amol Mandhane, Pratik Joshi, Anirudh Goyal, Matthew Tung, Andrew Brock, Hannah Sheahan, Vedant Misra, Cheng Li, Nemanja Rakićević, Mostafa Dehghani, Fangyu Liu, Sid Mittal, Junhyuk Oh, Seb Noury, Eren Sezener, Fantine Huot, Matthew Lamm, Nicola~De Cao, Charlie Chen, Sidharth Mudgal, Romina Stella, Kevin Brooks, Gautam Vasudevan, Chenxi Liu, Mainak Chain, Nivedita Melinkeri, Aaron Cohen, Venus Wang, Kristie Seymore, Sergey Zubkov, Rahul Goel, Summer Yue, Sai Krishnakumaran, Brian Albert, Nate Hurley, Motoki Sano, Anhad Mohananey, Jonah Joughin, Egor Filonov, Tomasz Kępa, Yomna Eldawy, Jiawern Lim, Rahul Rishi, Shirin Badiezadegan, Taylor Bos, Jerry Chang, Sanil Jain, Sri Gayatri~Sundara Padmanabhan, Subha Puttagunta, Kalpesh Krishna, Leslie Baker, Norbert Kalb, Vamsi Bedapudi, Adam Kurzrok, Shuntong Lei, Anthony Yu, Oren Litvin, Xiang Zhou, Zhichun Wu, Sam Sobell, Andrea Siciliano, Alan Papir, Robby Neale, Jonas Bragagnolo, Tej Toor, Tina Chen, Valentin Anklin, Feiran Wang, Richie Feng, Milad
  Gholami, Kevin Ling, Lijuan Liu, Jules Walter, Hamid Moghaddam, Arun Kishore, Jakub Adamek, Tyler Mercado, Jonathan Mallinson, Siddhinita Wandekar, Stephen Cagle, Eran Ofek, Guillermo Garrido, Clemens Lombriser, Maksim Mukha, Botu Sun, Hafeezul~Rahman Mohammad, Josip Matak, Yadi Qian, Vikas Peswani, Pawel Janus, Quan Yuan, Leif Schelin, Oana David, Ankur Garg, Yifan He, Oleksii Duzhyi, Anton Älgmyr, Timothée Lottaz, Qi~Li, Vikas Yadav, Luyao Xu, Alex Chinien, Rakesh Shivanna, Aleksandr Chuklin, Josie Li, Carrie Spadine, Travis Wolfe, Kareem Mohamed, Subhabrata Das, Zihang Dai, Kyle He, Daniel von Dincklage, Shyam Upadhyay, Akanksha Maurya, Luyan Chi, Sebastian Krause, Khalid Salama, Pam~G Rabinovitch, Pavan Kumar~Reddy M, Aarush Selvan, Mikhail Dektiarev, Golnaz Ghiasi, Erdem Guven, Himanshu Gupta, Boyi Liu, Deepak Sharma, Idan~Heimlich Shtacher, Shachi Paul, Oscar Akerlund, François-Xavier Aubet, Terry Huang, Chen Zhu, Eric Zhu, Elico Teixeira, Matthew Fritze, Francesco Bertolini, Liana-Eleonora
  Marinescu, Martin Bölle, Dominik Paulus, Khyatti Gupta, Tejasi Latkar, Max Chang, Jason Sanders, Roopa Wilson, Xuewei Wu, Yi-Xuan Tan, Lam~Nguyen Thiet, Tulsee Doshi, Sid Lall, Swaroop Mishra, Wanming Chen, Thang Luong, Seth Benjamin, Jasmine Lee, Ewa Andrejczuk, Dominik Rabiej, Vipul Ranjan, Krzysztof Styrc, Pengcheng Yin, Jon Simon, Malcolm~Rose Harriott, Mudit Bansal, Alexei Robsky, Geoff Bacon, David Greene, Daniil Mirylenka, Chen Zhou, Obaid Sarvana, Abhimanyu Goyal, Samuel Andermatt, Patrick Siegler, Ben Horn, Assaf Israel, Francesco Pongetti, Chih-Wei~"Louis" Chen, Marco Selvatici, Pedro Silva, Kathie Wang, Jackson Tolins, Kelvin Guu, Roey Yogev, Xiaochen Cai, Alessandro Agostini, Maulik Shah, Hung Nguyen, Noah~Ó Donnaile, Sébastien Pereira, Linda Friso, Adam Stambler, Adam Kurzrok, Chenkai Kuang, Yan Romanikhin, Mark Geller, ZJ~Yan, Kane Jang, Cheng-Chun Lee, Wojciech Fica, Eric Malmi, Qijun Tan, Dan Banica, Daniel Balle, Ryan Pham, Yanping Huang, Diana Avram, Hongzhi Shi, Jasjot Singh, Chris
  Hidey, Niharika Ahuja, Pranab Saxena, Dan Dooley, Srividya~Pranavi Potharaju, Eileen O'Neill, Anand Gokulchandran, Ryan Foley, Kai Zhao, Mike Dusenberry, Yuan Liu, Pulkit Mehta, Ragha Kotikalapudi, Chalence Safranek-Shrader, Andrew Goodman, Joshua Kessinger, Eran Globen, Prateek Kolhar, Chris Gorgolewski, Ali Ibrahim, Yang Song, Ali Eichenbaum, Thomas Brovelli, Sahitya Potluri, Preethi Lahoti, Cip Baetu, Ali Ghorbani, Charles Chen, Andy Crawford, Shalini Pal, Mukund Sridhar, Petru Gurita, Asier Mujika, Igor Petrovski, Pierre-Louis Cedoz, Chenmei Li, Shiyuan Chen, Niccolò~Dal Santo, Siddharth Goyal, Jitesh Punjabi, Karthik Kappaganthu, Chester Kwak, Pallavi LV, Sarmishta Velury, Himadri Choudhury, Jamie Hall, Premal Shah, Ricardo Figueira, Matt Thomas, Minjie Lu, Ting Zhou, Chintu Kumar, Thomas Jurdi, Sharat Chikkerur, Yenai Ma, Adams Yu, Soo Kwak, Victor Ähdel, Sujeevan Rajayogam, Travis Choma, Fei Liu, Aditya Barua, Colin Ji, Ji~Ho Park, Vincent Hellendoorn, Alex Bailey, Taylan Bilal, Huanjie Zhou,
  Mehrdad Khatir, Charles Sutton, Wojciech Rzadkowski, Fiona Macintosh, Konstantin Shagin, Paul Medina, Chen Liang, Jinjing Zhou, Pararth Shah, Yingying Bi, Attila Dankovics, Shipra Banga, Sabine Lehmann, Marissa Bredesen, Zifan Lin, John~Eric Hoffmann, Jonathan Lai, Raynald Chung, Kai Yang, Nihal Balani, Arthur Bražinskas, Andrei Sozanschi, Matthew Hayes, Héctor~Fernández Alcalde, Peter Makarov, Will Chen, Antonio Stella, Liselotte Snijders, Michael Mandl, Ante Kärrman, Paweł Nowak, Xinyi Wu, Alex Dyck, Krishnan Vaidyanathan, Raghavender R, Jessica Mallet, Mitch Rudominer, Eric Johnston, Sushil Mittal, Akhil Udathu, Janara Christensen, Vishal Verma, Zach Irving, Andreas Santucci, Gamaleldin Elsayed, Elnaz Davoodi, Marin Georgiev, Ian Tenney, Nan Hua, Geoffrey Cideron, Edouard Leurent, Mahmoud Alnahlawi, Ionut Georgescu, Nan Wei, Ivy Zheng, Dylan Scandinaro, Heinrich Jiang, Jasper Snoek, Mukund Sundararajan, Xuezhi Wang, Zack Ontiveros, Itay Karo, Jeremy Cole, Vinu Rajashekhar, Lara Tumeh, Eyal
  Ben-David, Rishub Jain, Jonathan Uesato, Romina Datta, Oskar Bunyan, Shimu Wu, John Zhang, Piotr Stanczyk, Ye~Zhang, David Steiner, Subhajit Naskar, Michael Azzam, Matthew Johnson, Adam Paszke, Chung-Cheng Chiu, Jaume~Sanchez Elias, Afroz Mohiuddin, Faizan Muhammad, Jin Miao, Andrew Lee, Nino Vieillard, Jane Park, Jiageng Zhang, Jeff Stanway, Drew Garmon, Abhijit Karmarkar, Zhe Dong, Jong Lee, Aviral Kumar, Luowei Zhou, Jonathan Evens, William Isaac, Geoffrey Irving, Edward Loper, Michael Fink, Isha Arkatkar, Nanxin Chen, Izhak Shafran, Ivan Petrychenko, Zhe Chen, Johnson Jia, Anselm Levskaya, Zhenkai Zhu, Peter Grabowski, Yu~Mao, Alberto Magni, Kaisheng Yao, Javier Snaider, Norman Casagrande, Evan Palmer, Paul Suganthan, Alfonso Castaño, Irene Giannoumis, Wooyeol Kim, Mikołaj Rybiński, Ashwin Sreevatsa, Jennifer Prendki, David Soergel, Adrian Goedeckemeyer, Willi Gierke, Mohsen Jafari, Meenu Gaba, Jeremy Wiesner, Diana~Gage Wright, Yawen Wei, Harsha Vashisht, Yana Kulizhskaya, Jay Hoover, Maigo Le,
  Lu~Li, Chimezie Iwuanyanwu, Lu~Liu, Kevin Ramirez, Andrey Khorlin, Albert Cui, Tian LIN, Marcus Wu, Ricardo Aguilar, Keith Pallo, Abhishek Chakladar, Ginger Perng, Elena~Allica Abellan, Mingyang Zhang, Ishita Dasgupta, Nate Kushman, Ivo Penchev, Alena Repina, Xihui Wu, Tom van~der Weide, Priya Ponnapalli, Caroline Kaplan, Jiri Simsa, Shuangfeng Li, Olivier Dousse, Fan Yang, Jeff Piper, Nathan Ie, Rama Pasumarthi, Nathan Lintz, Anitha Vijayakumar, Daniel Andor, Pedro Valenzuela, Minnie Lui, Cosmin Paduraru, Daiyi Peng, Katherine Lee, Shuyuan Zhang, Somer Greene, Duc~Dung Nguyen, Paula Kurylowicz, Cassidy Hardin, Lucas Dixon, Lili Janzer, Kiam Choo, Ziqiang Feng, Biao Zhang, Achintya Singhal, Dayou Du, Dan McKinnon, Natasha Antropova, Tolga Bolukbasi, Orgad Keller, David Reid, Daniel Finchelstein, Maria~Abi Raad, Remi Crocker, Peter Hawkins, Robert Dadashi, Colin Gaffney, Ken Franko, Anna Bulanova, Rémi Leblond, Shirley Chung, Harry Askham, Luis~C. Cobo, Kelvin Xu, Felix Fischer, Jun Xu, Christina Sorokin,
  Chris Alberti, Chu-Cheng Lin, Colin Evans, Alek Dimitriev, Hannah Forbes, Dylan Banarse, Zora Tung, Mark Omernick, Colton Bishop, Rachel Sterneck, Rohan Jain, Jiawei Xia, Ehsan Amid, Francesco Piccinno, Xingyu Wang, Praseem Banzal, Daniel~J. Mankowitz, Alex Polozov, Victoria Krakovna, Sasha Brown, MohammadHossein Bateni, Dennis Duan, Vlad Firoiu, Meghana Thotakuri, Tom Natan, Matthieu Geist, Ser tan Girgin, Hui Li, Jiayu Ye, Ofir Roval, Reiko Tojo, Michael Kwong, James Lee-Thorp, Christopher Yew, Danila Sinopalnikov, Sabela Ramos, John Mellor, Abhishek Sharma, Kathy Wu, David Miller, Nicolas Sonnerat, Denis Vnukov, Rory Greig, Jennifer Beattie, Emily Caveness, Libin Bai, Julian Eisenschlos, Alex Korchemniy, Tomy Tsai, Mimi Jasarevic, Weize Kong, Phuong Dao, Zeyu Zheng, Frederick Liu, Fan Yang, Rui Zhu, Tian~Huey Teh, Jason Sanmiya, Evgeny Gladchenko, Nejc Trdin, Daniel Toyama, Evan Rosen, Sasan Tavakkol, Linting Xue, Chen Elkind, Oliver Woodman, John Carpenter, George Papamakarios, Rupert Kemp, Sushant
  Kafle, Tanya Grunina, Rishika Sinha, Alice Talbert, Diane Wu, Denese Owusu-Afriyie, Cosmo Du, Chloe Thornton, Jordi Pont-Tuset, Pradyumna Narayana, Jing Li, Saaber Fatehi, John Wieting, Omar Ajmeri, Benigno Uria, Yeongil Ko, Laura Knight, Amélie Héliou, Ning Niu, Shane Gu, Chenxi Pang, Yeqing Li, Nir Levine, Ariel Stolovich, Rebeca Santamaria-Fernandez, Sonam Goenka, Wenny Yustalim, Robin Strudel, Ali Elqursh, Charlie Deck, Hyo Lee, Zonglin Li, Kyle Levin, Raphael Hoffmann, Dan Holtmann-Rice, Olivier Bachem, Sho Arora, Christy Koh, Soheil~Hassas Yeganeh, Siim Põder, Mukarram Tariq, Yanhua Sun, Lucian Ionita, Mojtaba Seyedhosseini, Pouya Tafti, Zhiyu Liu, Anmol Gulati, Jasmine Liu, Xinyu Ye, Bart Chrzaszcz, Lily Wang, Nikhil Sethi, Tianrun Li, Ben Brown, Shreya Singh, Wei Fan, Aaron Parisi, Joe Stanton, Vinod Koverkathu, Christopher~A. Choquette-Choo, Yunjie Li, TJ~Lu, Abe Ittycheriah, Prakash Shroff, Mani Varadarajan, Sanaz Bahargam, Rob Willoughby, David Gaddy, Guillaume Desjardins, Marco Cornero, Brona
  Robenek, Bhavishya Mittal, Ben Albrecht, Ashish Shenoy, Fedor Moiseev, Henrik Jacobsson, Alireza Ghaffarkhah, Morgane Rivière, Alanna Walton, Clément Crepy, Alicia Parrish, Zongwei Zhou, Clement Farabet, Carey Radebaugh, Praveen Srinivasan, Claudia van~der Salm, Andreas Fidjeland, Salvatore Scellato, Eri Latorre-Chimoto, Hanna Klimczak-Plucińska, David Bridson, Dario de~Cesare, Tom Hudson, Piermaria Mendolicchio, Lexi Walker, Alex Morris, Matthew Mauger, Alexey Guseynov, Alison Reid, Seth Odoom, Lucia Loher, Victor Cotruta, Madhavi Yenugula, Dominik Grewe, Anastasia Petrushkina, Tom Duerig, Antonio Sanchez, Steve Yadlowsky, Amy Shen, Amir Globerson, Lynette Webb, Sahil Dua, Dong Li, Surya Bhupatiraju, Dan Hurt, Haroon Qureshi, Ananth Agarwal, Tomer Shani, Matan Eyal, Anuj Khare, Shreyas~Rammohan Belle, Lei Wang, Chetan Tekur, Mihir~Sanjay Kale, Jinliang Wei, Ruoxin Sang, Brennan Saeta, Tyler Liechty, Yi~Sun, Yao Zhao, Stephan Lee, Pandu Nayak, Doug Fritz, Manish~Reddy Vuyyuru, John Aslanides, Nidhi Vyas,
  Martin Wicke, Xiao Ma, Evgenii Eltyshev, Nina Martin, Hardie Cate, James Manyika, Keyvan Amiri, Yelin Kim, Xi~Xiong, Kai Kang, Florian Luisier, Nilesh Tripuraneni, David Madras, Mandy Guo, Austin Waters, Oliver Wang, Joshua Ainslie, Jason Baldridge, Han Zhang, Garima Pruthi, Jakob Bauer, Feng Yang, Riham Mansour, Jason Gelman, Yang Xu, George Polovets, Ji~Liu, Honglong Cai, Warren Chen, XiangHai Sheng, Emily Xue, Sherjil Ozair, Christof Angermueller, Xiaowei Li, Anoop Sinha, Weiren Wang, Julia Wiesinger, Emmanouil Koukoumidis, Yuan Tian, Anand Iyer, Madhu Gurumurthy, Mark Goldenson, Parashar Shah, MK~Blake, Hongkun Yu, Anthony Urbanowicz, Jennimaria Palomaki, Chrisantha Fernando, Ken Durden, Harsh Mehta, Nikola Momchev, Elahe Rahimtoroghi, Maria Georgaki, Amit Raul, Sebastian Ruder, Morgan Redshaw, Jinhyuk Lee, Denny Zhou, Komal Jalan, Dinghua Li, Blake Hechtman, Parker Schuh, Milad Nasr, Kieran Milan, Vladimir Mikulik, Juliana Franco, Tim Green, Nam Nguyen, Joe Kelley, Aroma Mahendru, Andrea Hu, Joshua
  Howland, Ben Vargas, Jeffrey Hui, Kshitij Bansal, Vikram Rao, Rakesh Ghiya, Emma Wang, Ke~Ye, Jean~Michel Sarr, Melanie~Moranski Preston, Madeleine Elish, Steve Li, Aakash Kaku, Jigar Gupta, Ice Pasupat, Da-Cheng Juan, Milan Someswar, Tejvi M., Xinyun Chen, Aida Amini, Alex Fabrikant, Eric Chu, Xuanyi Dong, Amruta Muthal, Senaka Buthpitiya, Sarthak Jauhari, Nan Hua, Urvashi Khandelwal, Ayal Hitron, Jie Ren, Larissa Rinaldi, Shahar Drath, Avigail Dabush, Nan-Jiang Jiang, Harshal Godhia, Uli Sachs, Anthony Chen, Yicheng Fan, Hagai Taitelbaum, Hila Noga, Zhuyun Dai, James Wang, Chen Liang, Jenny Hamer, Chun-Sung Ferng, Chenel Elkind, Aviel Atias, Paulina Lee, Vít Listík, Mathias Carlen, Jan van~de Kerkhof, Marcin Pikus, Krunoslav Zaher, Paul Müller, Sasha Zykova, Richard Stefanec, Vitaly Gatsko, Christoph Hirnschall, Ashwin Sethi, Xingyu~Federico Xu, Chetan Ahuja, Beth Tsai, Anca Stefanoiu, Bo~Feng, Keshav Dhandhania, Manish Katyal, Akshay Gupta, Atharva Parulekar, Divya Pitta, Jing Zhao, Vivaan Bhatia,
  Yashodha Bhavnani, Omar Alhadlaq, Xiaolin Li, Peter Danenberg, Dennis Tu, Alex Pine, Vera Filippova, Abhipso Ghosh, Ben Limonchik, Bhargava Urala, Chaitanya~Krishna Lanka, Derik Clive, Yi~Sun, Edward Li, Hao Wu, Kevin Hongtongsak, Ianna Li, Kalind Thakkar, Kuanysh Omarov, Kushal Majmundar, Michael Alverson, Michael Kucharski, Mohak Patel, Mudit Jain, Maksim Zabelin, Paolo Pelagatti, Rohan Kohli, Saurabh Kumar, Joseph Kim, Swetha Sankar, Vineet Shah, Lakshmi Ramachandruni, Xiangkai Zeng, Ben Bariach, Laura Weidinger, Tu~Vu, Alek Andreev, Antoine He, Kevin Hui, Sheleem Kashem, Amar Subramanya, Sissie Hsiao, Demis Hassabis, Koray Kavukcuoglu, Adam Sadovsky, Quoc Le, Trevor Strohman, Yonghui Wu, Slav Petrov, Jeffrey Dean, and Oriol Vinyals. 2024.
\newblock \href {https://arxiv.org/abs/2312.11805} {Gemini: A family of highly capable multimodal models}.
\newblock \emph{Preprint}, arXiv:2312.11805.

\bibitem[{Gerz et~al.(2016)Gerz, Vuli{\'c}, Hill, Reichart, and Korhonen}]{gerz-etal-2016-simverb}
Daniela Gerz, Ivan Vuli{\'c}, Felix Hill, Roi Reichart, and Anna Korhonen. 2016.
\newblock \href {https://doi.org/10.18653/v1/D16-1235} {{S}im{V}erb-3500: A large-scale evaluation set of verb similarity}.
\newblock In \emph{Proceedings of the 2016 Conference on Empirical Methods in Natural Language Processing}, pages 2173--2182, Austin, Texas. Association for Computational Linguistics.

\bibitem[{Herbelot and Baroni(2017)}]{herbelot2017high}
Aur{\'e}lie Herbelot and Marco Baroni. 2017.
\newblock High-risk learning: acquiring new word vectors from tiny data.
\newblock \emph{arXiv preprint arXiv:1707.06556}.

\bibitem[{Hill et~al.(2015)Hill, Reichart, and Korhonen}]{hill-etal-2015-simlex}
Felix Hill, Roi Reichart, and Anna Korhonen. 2015.
\newblock \href {https://doi.org/10.1162/COLI_a_00237} {{S}im{L}ex-999: Evaluating semantic models with (genuine) similarity estimation}.
\newblock \emph{Computational Linguistics}, 41(4):665--695.

\bibitem[{Jacobs et~al.(2022)Jacobs, Hubbard, and Federmeier}]{jacobs-etal-2022-masked}
Cassandra~L. Jacobs, Ryan~J. Hubbard, and Kara~D. Federmeier. 2022.
\newblock \href {https://aclanthology.org/2022.scil-1.22} {Masked language models directly encode linguistic uncertainty}.
\newblock In \emph{Proceedings of the Society for Computation in Linguistics 2022}, pages 225--228, online. Association for Computational Linguistics.

\bibitem[{Kapelner et~al.(2018)Kapelner, Soterwood, Nessaiver, and Adlof}]{Kapelner_2018}
Adam Kapelner, Jeanine Soterwood, Shalev Nessaiver, and Suzanne Adlof. 2018.
\newblock \href {https://doi.org/10.1109/TLT.2018.2789900} {Predicting contextual informativeness for vocabulary learning}.
\newblock \emph{IEEE Transactions on Learning Technologies}, 11(1):13–26.

\bibitem[{Liu et~al.(2019{\natexlab{a}})Liu, Ott, Goyal, Du, Joshi, Chen, Levy, Lewis, Zettlemoyer, and Stoyanov}]{Liu2019RoBERTaAR}
Yinhan Liu, Myle Ott, Naman Goyal, Jingfei Du, Mandar Joshi, Danqi Chen, Omer Levy, Mike Lewis, Luke Zettlemoyer, and Veselin Stoyanov. 2019{\natexlab{a}}.
\newblock {RoBERTa}: {A} robustly optimized {BERT} pretraining approach.
\newblock \emph{ArXiv}, abs/1907.11692.

\bibitem[{Liu et~al.(2019{\natexlab{b}})Liu, Ott, Goyal, Du, Joshi, Chen, Levy, Lewis, Zettlemoyer, and Stoyanov}]{roberta}
Yinhan Liu, Myle Ott, Naman Goyal, Jingfei Du, Mandar Joshi, Danqi Chen, Omer Levy, Mike Lewis, Luke Zettlemoyer, and Veselin Stoyanov. 2019{\natexlab{b}}.
\newblock \href {https://arxiv.org/abs/1907.11692} {Roberta: A robustly optimized bert pretraining approach}.
\newblock \emph{Preprint}, arXiv:1907.11692.

\bibitem[{Nagy and Anderson(1984)}]{Nagy_Anderson_1984}
William~E. Nagy and Richard~C. Anderson. 1984.
\newblock \href {https://doi.org/10.2307/747823} {How many words are there in printed school english?}
\newblock \emph{Reading Research Quarterly}, 19(3):304–330.

\bibitem[{Nam et~al.(2022)Nam, Jurgens, and Collins-Thompson}]{Nam2022}
Sungjin Nam, David Jurgens, and Kevyn Collins-Thompson. 2022.
\newblock \href {http://arxiv.org/abs/2204.09885} {An attention-based model for predicting contextual informativeness and curriculum learning applications}.
\newblock (arXiv:2204.09885).
\newblock ArXiv:2204.09885 [cs].

\bibitem[{Rapaport(2005)}]{rapaport2005defense}
William~J Rapaport. 2005.
\newblock In defense of contextual vocabulary acquisition: How to do things with words in context.
\newblock In \emph{International and interdisciplinary conference on modeling and using context}, pages 396--409. Springer.

\bibitem[{Saedi et~al.(2018)Saedi, Branco, António~Rodrigues, and Silva}]{Saedi2018}
Chakaveh Saedi, António Branco, João António~Rodrigues, and João Silva. 2018.
\newblock \href {https://doi.org/10.18653/v1/W18-3016} {Wordnet embeddings}.
\newblock page 122–131, Melbourne, Australia. Association for Computational Linguistics.

\bibitem[{Schick and Sch{\"u}tze(2019)}]{schick2019learning}
Timo Schick and Hinrich Sch{\"u}tze. 2019.
\newblock Learning semantic representations for novel words: Leveraging both form and context.
\newblock In \emph{Proceedings of the AAAI Conference on Artificial Intelligence}, volume~33, pages 6965--6973.

\bibitem[{Speer et~al.(2017)Speer, Chin, and Havasi}]{speer2017conceptnet}
Robyn Speer, Joshua Chin, and Catherine Havasi. 2017.
\newblock \href {http://aaai.org/ocs/index.php/AAAI/AAAI17/paper/view/14972} {{ConceptNet} 5.5: An open multilingual graph of general knowledge}.
\newblock pages 4444--4451.

\bibitem[{Taylor(1953)}]{taylor1953cloze}
Wilson~L Taylor. 1953.
\newblock “cloze procedure”: A new tool for measuring readability.
\newblock \emph{Journalism quarterly}, 30(4):415--433.

\bibitem[{Toshevska et~al.(2020)Toshevska, Stojanovska, and Kalajdjieski}]{toshevska2020comparative}
Martina Toshevska, Frosina Stojanovska, and Jovan Kalajdjieski. 2020.
\newblock Comparative analysis of word embeddings for capturing word similarities.
\newblock \emph{arXiv preprint arXiv:2005.03812}.

\bibitem[{Valentini et~al.(2023)Valentini, Weber, Salcido, Wright, Colunga, and von~der Wense}]{valentini}
Maria Valentini, Jennifer Weber, Jesus Salcido, T{\'e}a Wright, Eliana Colunga, and Katharina von~der Wense. 2023.
\newblock \href {https://doi.org/10.18653/v1/2023.emnlp-main.218} {On the automatic generation and simplification of children{'}s stories}.
\newblock In \emph{Proceedings of the 2023 Conference on Empirical Methods in Natural Language Processing}, pages 3588--3598, Singapore. Association for Computational Linguistics.

\bibitem[{Walker et~al.(1994)Walker, Greenwood, Hart, and Carta}]{Walker_Greenwood_Hart_Carta_1994}
Dale Walker, Charles Greenwood, Betty Hart, and Judith Carta. 1994.
\newblock \href {https://doi.org/10.2307/1131404} {Prediction of school outcomes based on early language production and socioeconomic factors}.
\newblock \emph{Child Development}, 65(2):606–621.

\bibitem[{Webb(2008)}]{webb}
Stuart Webb. 2008.
\newblock \href {https://files.eric.ed.gov/fulltext/EJ815123.pdf} {The effects of context on incidental vocabulary learning}.
\newblock \emph{Reading in a Foreign Language}, 20:232--245.

\bibitem[{Wolf et~al.(2020)Wolf, Debut, Sanh, Chaumond, Delangue, Moi, Cistac, Rault, Louf, Funtowicz, Davison, Shleifer, von Platen, Ma, Jernite, Plu, Xu, Scao, Gugger, Drame, Lhoest, and Rush}]{wolf-etal-2020-transformers}
Thomas Wolf, Lysandre Debut, Victor Sanh, Julien Chaumond, Clement Delangue, Anthony Moi, Pierric Cistac, Tim Rault, Rémi Louf, Morgan Funtowicz, Joe Davison, Sam Shleifer, Patrick von Platen, Clara Ma, Yacine Jernite, Julien Plu, Canwen Xu, Teven~Le Scao, Sylvain Gugger, Mariama Drame, Quentin Lhoest, and Alexander~M. Rush. 2020.
\newblock \href {https://www.aclweb.org/anthology/2020.emnlp-demos.6} {Transformers: State-of-the-art natural language processing}.
\newblock In \emph{Proceedings of the 2020 Conference on Empirical Methods in Natural Language Processing: System Demonstrations}, pages 38--45, Online. Association for Computational Linguistics.

\end{thebibliography}

\pagebreak
 
\appendix

\section{Selection of Word Embeddings}
\label{appendix:wordembeddings}

\subsection{Word Embeddings}
ConceptNet Numberbatch 19.08 English word embeddings \cite{speer2017conceptnet} are considered state-of-the-art and have been shown to correlate best with human discernment of similarity between word pairs on three gold-standard similarity datasets: SimLex-999 \cite{hill-etal-2015-simlex}, SimVerb-3500 \cite{gerz-etal-2016-simverb}, and WordSimilarity-353 \cite{finkelstein2001placing}. Cosine similarity between two words using ConceptNet Numberbatch embeddings has the highest correlation with gold standard scores using Spearman's Rho, Pearson's R, and Kendall's Tau correlation \cite{toshevska2020comparative}. We verify these results on the top performing embeddings from \citet{toshevska2020comparative} for SimLex-999 and WordSimilarity-353 in Table \ref{table3}.

\begin{table}[h]
\small 
\centering
\begin{tabular}{l|cc|cc}
\toprule
\multicolumn{5}{c}{\textbf{SimLex-999}} \\
\midrule
& \multicolumn{2}{c|}{\textbf{$r$}} & \multicolumn{2}{c}{\textbf{$\rho$}} \\
\midrule
\textbf{Word2Vec(GoogleNews 300)} & \multicolumn{2}{c|}{0.4539} & \multicolumn{2}{c}{0.4420} \\
\textbf{LexVec(CommonCrawl 300)} & \multicolumn{2}{c|}{0.4542} & \multicolumn{2}{c}{0.4442} \\
\textbf{ConceptNet Numberbatch 19.08} & \multicolumn{2}{c|}{\textbf{0.6458}} & \multicolumn{2}{c}{\textbf{0.6268}} \\
\midrule
\multicolumn{5}{c}{\textbf{WordSimilarity-353}} \\
\midrule
& \multicolumn{2}{c|}{\textbf{$r$}} & \multicolumn{2}{c}{\textbf{$\rho$}} \\
\midrule
\textbf{Word2Vec(GoogleNews 300)} & \multicolumn{2}{c|}{0.6411} & \multicolumn{2}{c}{0.6833} \\
\textbf{LexVec(CommonCrawl 300)} & \multicolumn{2}{c|}{0.6845} & \multicolumn{2}{c}{0.7189} \\
\textbf{ConceptNet Numberbatch 19.08} & \multicolumn{2}{c|}{\textbf{0.7534}} & \multicolumn{2}{c}{\textbf{0.8149}} \\
\bottomrule
\end{tabular}
\caption{Verification of word embedding performance against similarity gold standard evaluation datasets. $r$ indicates Pearson's $r$ and $\rho$ indicates Spearman's $\rho$.}
\label{table3}
\end{table}

\section{Computational Experiment Details}
\label{appendix:computationalexperimentdetails}

\subsection{Existing Packages}
The packages and versions we use for our implementation include: transformers 4.37.2, pandas 2.2.0, numpy 1.26.4, NLTK 3.8.1, gensim 4.3.2, scikit-learn 1.4.0, and scipy 1.12.0.
\subsection{Model Parameters}
For our implementation of RoBERTa, we use roberta-base, which has 125M parameters. 
\subsection{Model Hyperparameters}
Hyperparameters for RoBERTaForMaskedLM:\\
"attention$\_$probs$\_$dropout$\_$prob": 0.1, \\
"bos$\_$token$\_$id": 0, \\
"classifier$\_$dropout": null, \\
"eos$\_$token$\_$id": 2, \\
"hidden$\_$act": "gelu", \\
"hidden$\_$dropout$\_$prob": 0.1, \\
"hidden$\_$size": 768, \\
"initializer$\_$range": 0.02, \\
"intermediate$\_$size": 3072, \\
"layer$\_$norm$\_$eps": 1e-05, \\
"max$\_$position$\_$embeddings": 514, \\
"model$\_$type": "roberta", \\
"num$\_$attention$\_$heads": 12, \\
"num$\_$hidden$\_$layers": 12, \\
"pad$\_$token$\_$id": 1, \\
"position$\_$embedding$\_$type": "absolute", \\
"transformers$\_$version": "4.37.2", \\
"type$\_$vocab$\_$size": 1, \\
"use$\_$cache": true, \\
"vocab$\_$size": 50265 \\\\

\section{Semantic Similarity Baselines}
\label{appendix:contextbaselines}

\subsection{Semantic Similarity Baselines}
In the main results tables, we utilize the best performing semantic similarity baselines of three lengths tested for Context Window and three thresholds tested for Number Relevant Words.
\begin{table}[h]
\small 
\centering
\begin{tabular}{l|cc|cc|cc}
\multicolumn{7}{c}{\textbf{Baseline Comparison}}
\\ 
\midrule
& \multicolumn{2}{c|}{\textbf{$r$}}                                   
& \multicolumn{2}{c|}{\textbf{$\rho$}}                                   
& \multicolumn{2}{c}{\textbf{RMSE}}                                                  
\\ 
\midrule
\textbf{Context Similarity}                                
& \multicolumn{2}{c|}{0.2858}                                     
& \multicolumn{2}{c|}{0.2890}                                     
& \multicolumn{2}{c}{0.3078}                                                         
\\ 
\textbf{Context Window(1 word)}                                
& \multicolumn{2}{c|}{0.1587}                                     
& \multicolumn{2}{c|}{0.2036}                                     
& \multicolumn{2}{c}{0.3583}                                         
\\
\textbf{Context Window(3 words)}                                
& \multicolumn{2}{c|}{0.2397}                                     
& \multicolumn{2}{c|}{0.2815}                                     
& \multicolumn{2}{c}{\textbf{0.2918}}                                         
\\
\textbf{Context Window(5 words)}                                
& \multicolumn{2}{c|}{0.2772}                                     
& \multicolumn{2}{c|}{0.3134}                                     
& \multicolumn{2}{c}{0.2921}                                         
\\

\textbf{Num Related Words(0.3)}                                
& \multicolumn{2}{c|}{\textbf{0.3239}}                                     
& \multicolumn{2}{c|}{\textbf{0.3534}}                                     
& \multicolumn{2}{c}{0.4120}    
\\

\textbf{Num Related Words(0.4)}                                
& \multicolumn{2}{c|}{0.3056}                                     
& \multicolumn{2}{c|}{0.3523}                                     
& \multicolumn{2}{c}{0.4387}                                                                            
\\

\textbf{Num Related Words(0.5)}                                
& \multicolumn{2}{c|}{0.2451}                                     
& \multicolumn{2}{c|}{0.2483}                                     
& \multicolumn{2}{c}{0.4683}                                                                            
\\ \bottomrule
\end{tabular}
\caption{Comparison of cosine similarity-based context baselines. $r$ indicates Pearson's $r$ and $\rho$ indicates Spearman's $\rho$.}
\label{table4}
\end{table}

\section{Nam et al. Model Performance}
\label{appendix:likert}

\subsection{Nam et al. Performance and Annotation Comparison}

\begin{table}[H]
\small 
\centering
\begin{tabular}{l|cc|cc|cc}

\toprule
\multicolumn{7}{c}{\textbf{Kapelner et al. data, Likert scale gold standard}}
\\ 
\midrule
& \multicolumn{2}{c|}{\textbf{$r$}}                                   
& \multicolumn{2}{c|}{\textbf{$\rho$}}                                   
& \multicolumn{2}{c}{\textbf{RMSE}}                                                  
\\ 
\midrule
\textbf{Nam et al.}                                
& \multicolumn{2}{c|}{0.6691}                                         
& \multicolumn{2}{c|}{0.6286}                                         
& \multicolumn{2}{c}{0.2026}                                         
\\ 
\textbf{Nam et al. + WordNet}                         
& \multicolumn{2}{c|}{\textbf{0.6815}}                                  
& \multicolumn{2}{c|}{\textbf{0.6496}}                                        
& \multicolumn{2}{c}{\textbf{0.1660}}                      
\\
\midrule
\multicolumn{7}{c}{\textbf{Kapelner et al. data, cloze gold standard}}
\\ 
\midrule
& \multicolumn{2}{c|}{\textbf{$r$}}                                   
& \multicolumn{2}{c|}{\textbf{$\rho$}}                                   
& \multicolumn{2}{c}{\textbf{RMSE}}                                                 
\\ 
\midrule
\textbf{Nam et al.}                                
& \multicolumn{2}{c|}{0.3217}                                         
& \multicolumn{2}{c|}{0.3545}                                         
& \multicolumn{2}{c}{0.3971}                                         
\\ 
\textbf{Nam et al. + WordNet}                         
& \multicolumn{2}{c|}{\textbf{0.3230}}                                      
& \multicolumn{2}{c|}{\textbf{0.3660}}                                        
& \multicolumn{2}{c}{\textbf{0.3540}}                      
\\
\midrule

\multicolumn{7}{c}{\textbf{Child-Directed data, cloze gold standard}}
\\ 
\midrule
& \multicolumn{2}{c|}{\textbf{$r$}}                                   
& \multicolumn{2}{c|}{\textbf{$\rho$}}                                   
& \multicolumn{2}{c}{\textbf{RMSE}}                                                 \\ 
\midrule
\textbf{Nam et al.}                                
& \multicolumn{2}{c|}{0.0505}                                         
& \multicolumn{2}{c|}{0.0525}                                         
& \multicolumn{2}{c}{\textbf{0.3165}}                                         
\\

\textbf{Nam et al. + WordNet}                         
& \multicolumn{2}{c|}{\textbf{0.0623}}                                        
& \multicolumn{2}{c|}{\textbf{0.0574}}                                        
& \multicolumn{2}{c}{0.3166}                                                                      
\\ \bottomrule
\end{tabular}
\caption{Results of the Nam et al. model on different annotation schemas and datasets. $r$ indicates Pearson's $r$ and $\rho$ indicates Spearman's $\rho$.}
\label{table5}
\end{table}

\section{Human Annotation Details}
\label{appendix:annotators}

\subsection{Annotator Instructions}
The annotators received the following instructions prior to beginning the survey where we collected annotations. "In the following survey you will see a set of 60 children’s stories, one per page. Each story has FIVE words blanked out, labeled 1-5. Your job is to try to guess each of the FIVE words from context.

Each word could be a noun (person, place, or thing), a verb (action word), or an adjective (descriptive word). Each word may occur more than one time, so please read the whole story before making your best guess for each word. Additionally, any word may appear in different forms throughout the story. For example, the noun apple could appear in the story sometimes as apple or as apples. Similarly verbs might appear in different forms – walk, walking, walked, or walks – then you would write “walking” for it. Even if you think of more than one potential word for each missing word, write down the one you think fits best.

At the end of each story, you will also be asked to evaluate whether you think the story would be appropriate for a preschooler (3-5-years-old).

If you need to take a break, you can do that. Clicking the same link will take you to the last page you completed so you can continue the survey."

\end{document}